\documentclass[letter]{article}
\usepackage{spconf,amsmath,graphicx}

\usepackage[utf8]{inputenc}
\usepackage{hyperref}       
\usepackage{url}            
\usepackage{booktabs}       
\usepackage{amsfonts}       
\usepackage{nicefrac}       
\usepackage{microtype}      

\usepackage{color,graphicx,amssymb,amsmath,amsthm}

\def\argmax{\mathop{\mbox{argmax}}}
\def\argmin{\mathop{\mbox{argmin}}}
\def\range{\mathop{\mbox{range}}}
\def\st{\mathop{\mbox{subject to}}}
\def \bx{\boldsymbol{x}}

\def \by{\boldsymbol{y}}

\def \bd{\boldsymbol{d}}
\def \bz{\boldsymbol{z}}
\def \bw{\boldsymbol{w}}
\def \bu{\boldsymbol{u}}
\def \bv{\boldsymbol{v}}

\def \bs{\boldsymbol{s}}
\def \bq{\boldsymbol{q}}

\def \bA{\boldsymbol{A}}
\def \bB{\boldsymbol{B}}

\def \bG{\boldsymbol{G}}
\def \bD{\boldsymbol{D}}
\def \bF{\boldsymbol{F}}
\def \bI{\boldsymbol{I}}
\def \bP{\boldsymbol{P}}
\def \bQ{\boldsymbol{Q}}

\def \bgamma{\mbox{\boldmath $\gamma$}}
\def \bDelta{\mbox{\boldmath $\Delta$}}
\def \bDelta{\mbox{\boldmath $\Delta$}}

\def \btheta{\mbox{\boldmath $\theta$}}

\title{SYNTHESIS VERSUS ANALYSIS IN PATCH-BASED IMAGE PRIORS}

\name{Mário A. T. Figueiredo}
\address{Instituto de Telecomunica\c{c}\~{o}es and Instituto Superior Técnico, 
 University of Lisbon, Portugal\\
 \texttt{mario.figueiredo@tecnico.ulisboa.pt} \\
} 

\ninept

\begin{document}

\maketitle

\begin{abstract}
In global models/priors (for example, using wavelet frames), there is a well known analysis vs synthesis dichotomy in the way signal/image priors are formulated.  In patch-based image models/priors, this dichotomy is also present in the choice of how each patch is modeled. This paper shows that there is another analysis vs synthesis dichotomy, in terms of how the whole image is related to the patches, and that all existing patch-based formulations that provide a global image prior belong to the analysis category. We then propose a synthesis formulation, where the image is explicitly modeled as being synthesized by additively combining a collection of independent patches. We formally establish that these analysis and synthesis formulations are not equivalent in general and that both formulations are compatible with analysis and synthesis formulations at the patch level.
Finally, we present an instance of the \textit{alternating direction method of multipliers} (ADMM) that can be used to perform image denoising under the proposed synthesis formulation, showing its computational feasibility. Rather than showing the superiority of the synthesis or analysis formulations, the contributions of this paper is to establish the existence of both alternatives, thus closing the corresponding gap in the field of patch-based image processing. 
\end{abstract}

\section{Introduction}\label{sec:intro}
The use of patches in image processing can be seen as an instance of the ``divide and conquer" principle: since it is admittedly very difficult to formulate a global prior/model for images, patch-based approaches use priors/models for patches (rather than whole images), the combination of which yields the desired image prior/model. To keep the discussion and formulation at their essential and focus on the image modelling aspects, we will concentrate on image denoising, arguably the quintessential image processing problem. Nevertheless, much of what will be presented below can be easily extended (at least in principle) to more general inverse problems.

There are basically two approaches to patch-based image denoising. In earlier methods  \cite{Buades}, \cite{Dabov}, \cite{EladAharon}, patches are extracted from the noisy image, then processed/denoised independently (or maybe even collaboratively, as in BM3D \cite{Dabov}), and finally returned to their original locations. Since the patches overlap (to avoid blocking artifacts), there are several  estimates of each pixel, which are combined by some form of averaging (unweighted or weighted). This approach is also used in the nonlocal Bayesian method \cite{LeBrun}, and in methods based on Gaussian mixtures \cite{Teodoro}. For a comprehensive review of these and related methods, see \cite{LeBrun}. Arguably, a conceptual flaw of these methods is that they obtain patch estimates without explicitly taking into account that these will subsequently be combined additively. As a consequence, although some of these methods achieve state-of-the-art results, they do not explicitly provide a global image prior/model. 

A more recent class of approaches does build global image models that are based on a function computed from image patches, but does not treat them as independent by explicitly taking into account that they are overlapping patches of the same image; this approach was initiated with the \textit{expected patch log-likelihood} (EPLL) \cite{ZoranWeiss}, and is adopted by most of the recent work \cite{Chen}, \cite{SulamElad}. These methods do not have the conceptual flaw pointed out in the previous paragraph and  provide a coherent global image model. 

The analysis vs synthesis dichotomy in global image models/priors  (\textit{e.g.}, based on wavelet frames, or total variation) has been first formalized in \cite{EladMilanfar}, and further studied in \cite{Selesnick}; more recently, it has been ported to patch-wise models \cite{Chen}. To the best of our knowledge, this dichotomy has not been pointed out before concerning the way in which  patch-level models/priors are used to build a global image models; that is precisely the central contribution of this paper.

In the synthesis vs analysis dichotomy, the EPLL-type class of patch-based models can be seen as an analysis method (as explained below in detail). This paper shows that there exists the synthesis counterpart of EPLL; in other words, that the synthesis vs analysis dichotomy is also present in the way the whole image and the patches are related. We also show that the two formulations are not, in general, equivalent.

The remaining sections of this paper are organized as follows. After reviewing the classical analysis/synthesis dichotomy in Section \ref{sec:a_vs_s}, we shown in Section \ref{sec:patch_a} that the classical patch-based methods follow an analysis formulation. Section \ref{sec:patch_s} then introduces a synthesis patch-based formulation, and its relationship with the analysis counterpart is established in Section \ref{sec:rel}. In Section \ref{sec:admm}, we present an ADMM algorithm to efficiently perform image denoising under the proposed synthesis formulation. Finally, Section \ref{sec:conclusion} concludes the paper by referring to future work directions.

\section{Analysis vs Synthesis in Image Denoising}~\label{sec:a_vs_s}
Before addressing patch-based models, we briefly review the analysis and synthesis global formulations of image denoising \cite{EladMilanfar}, where the goal is to estimate an unknown image $\bx  \in \mathbb{R}^N$ ($N$ is the total number of pixels in $\bx$, which is a vectorized version of the corresponding $\sqrt{N}\times \sqrt{N}$ image) from a noisy version thereof
\begin{equation}
\by = \bx + \bw,
\end{equation}  
where $\bw$ is a sample of a white Gaussian noise field of zero mean and (known) variance $\sigma^2$, that is, $p(\by|\bx) = \mathcal{N}(\by;\bx,\sigma^2 \bI)$. 

The classical approach to estimate $\bx$ from  $\by$ is to adopt (or learn) a prior $p_X$ for the unknown $\bx$ and seek a maximizer of the posterior density  (a {\it maximum  a posteriori}--MAP--estimate)
\begin{equation}
\hat{\bx}_{\mbox{\scriptsize MAP}}  \in \argmax_{\bx} p(\bx | \by) = \argmin_{\bx} \frac{1}{2\sigma^2} \|\by - \bx\|_2^2 - \log p_X(\bx),
\end{equation}
where $\frac{1}{2\sigma^2} \|\by - \bx\|_2^2 = -\log p(\by|\bx)$, up to an irrelevant constant. The analysis and synthesis formulations build priors for $\bx$ as follows.
\begin{description}
	\item[Analysis:] in this formulation, the prior $p_X$ takes the form 
	\begin{equation}
	-\log p_X(\bx) = \varphi (\bF \bx) + b,
	\end{equation}
	where $b$ is a constant and $\bF$ a linear (analysis) operator. In this case, the MAP estimate takes the form
    \begin{equation}
   \hat{\bx}_{\mbox{\scriptsize MAP-A}}  \in \argmax_{\bx} p(\bx | \by) = \argmin_{\bx} \frac{1}{2\sigma^2} \|\by - \bx\|_2^2 +  \varphi (\bF \bx), \label{eq:analysis_general}
    \end{equation}
    where MAP-A stands for \textit{MAP-analysis}.
    \item[Synthesis:] here, the starting point is to assume that $\bx$ is linearly synthesized/represented according to $\bx = \bG \btheta$, where $\btheta$ is a vector of coefficients, and the prior is formulated on $\btheta$ rather than directly on $\bx$; once an estimate $\hat\btheta$ is obtained, the corresponding estimate of $\bx$ is simply $\hat{\bx} = \bG\hat{\btheta}$. In summary,
    \begin{equation}
    \hat{\bx}_{\mbox{\scriptsize MAP-S}} = \bG \hat{\btheta}, \; \mbox{where}\;  \hat{\btheta} \in \argmin_{\btheta} \frac{1}{2\sigma^2} \|\by -\bG \btheta\|_2^2 +\zeta ( \btheta),
    \end{equation}
    and $\zeta(\btheta) = -\log_{\Theta}(\btheta)$ is a negative log-prior on $\btheta$.
\end{description}
One of the main distinguishing features of analysis and synthesis formulations is that in the former the object of the estimation procedure is the image itself, whereas in the latter, one estimates a representation from which the image estimate is synthesized.

\section{Patch-Analysis Formulation}\label{sec:patch_a}
\subsection{Expected Patch Log-likelihood}
A central tool in most patch-based approaches is a collection of operators $\{\bP_m, \, m=1,..., M\}$ that extracts $M$ patches from a given image of  size $\sqrt{N}\times \sqrt{N}$; each $\bP_m$ can be seen as a binary matrix with size $n \times N$, where $n$ is the total number of pixels in each patch (assumed square, of size $\sqrt{n}\times \sqrt{n}$). The standard way of formulating a patch-based prior is by writing
\begin{equation}
p_X ( \bx ) = \frac{1}{Z}\prod_{m=1}^M f(\bP_m \bx) \label{eq:PoE},
\end{equation}
where $f: \mathbb{R}^n \rightarrow \mathbb{R}_+$ is a function expressing the patch-wise prior distribution and $Z$ is a normalizing constant. Function $f$ may itself be a \textit{probability density function} (pdf), \textit{e.g.}, a GMM,  in which case this prior is an instance of a so-called {\it product of experts} (PoE \cite{PoE}). However, $f$ does not need to be a pdf, as long as it takes non-negative values; in fact, \eqref{eq:PoE} can also be seen as a \textit{factor graph} model, where each factor corresponds to a patch and the all factors have the same function $f$ \cite{WainwrightJordan}. Moreover, this prior is also equivalent to the formulation known as EPLL (\textit{expected patch log-likleihood} \cite{ZoranWeiss}), although EPLL was not originally interpreted as a prior.

Given a noisy image $\by$, a MAP estimate of $\bx$  is given by 
\begin{equation}
\hat{\bx}_{\mbox{\scriptsize MAP-A}} \in \arg\min_{\bx} \frac{1}{2\sigma^2} \|\by - \bx\|_2^2 - \sum _{m=1}^M \log f(\bP_m \bx). \label{eq:analysis}
\end{equation}

\subsection{Half Quadratic Splitting and ADMM}
To tackle the large-scale optimization problem in \eqref{eq:analysis}, the so-called \textit{half quadratic splitting} strategy replaces it with
\begin{eqnarray}
\hat{\bx}_{\mbox{\scriptsize MAP-A}} & \displaystyle \!\!\! \in  \arg \min_{\bx} \min_{\bv} & \!\!\! \frac{1}{\sigma^2} \|\by - \bx\|_2^2  + \beta \sum _{m=1}^M \|\bv_m - \bP_m\bx\|_2^2 \nonumber \\ &  & - 2 \sum _{m=1}^M \log f(\bv_m),\label{eq:EPLL}
\end{eqnarray}
where $\bv = (\bv_m,\; m=1,...,M)$, which obviously becomes equivalent to \eqref{eq:analysis} as $\beta\rightarrow \infty$ \cite{ZoranWeiss}. The optimization problem in \eqref{eq:EPLL} is tackled by alternating between minimizing with respect to $\bv$ and $\bx$, while slowly increasing $\beta$.  Other strategies for setting $\beta$ have also been proposed \cite{ZoranWeiss}.

An obvious alternative (as recently mentioned in \cite{Papyan}) is to reformulate \eqref{eq:analysis} as a constrained problem
\begin{eqnarray}
\hat{x}_{\mbox{\scriptsize MAP-A}} & = & \arg \min_{\bx} \frac{1}{2\sigma^2} \|\by - \bx\|_2^2 - \sum _{m=1}^M \log f( \bv_m ) \label{eq:analysis_admm1}\\
& & \mbox{subject to} \; \bv_m = \bP_m \bx, \;\;\mbox{for $m=1,...,M$}\nonumber
\end{eqnarray}
and tackle it with ADMM ({\it alternating direction method of multipliers}) \cite{Boyd}. 
Of course, convergence of ADMM for this problem can only be guaranteed if 
the negative log factors $- \log f$ are convex; this is not the case if $f$ is a GMM, but it is true if $- \log f$ is an $\ell_r$ norm or the $r$-th power thereof, \textit{e.g.},  $- \log f(\bv_m) = \|\bv_m\|_r^r$, with $r\geq 1$.

Examining \eqref{eq:EPLL} (with $\beta\rightarrow\infty$) or \eqref{eq:analysis_admm1} reveals that this formulation seeks a consensus among the patches, in the sense that the several replicates of each pixels that exist in different patches are forced to agree on a common value for that pixel. 
In other words, the clean patches are not modelled as additively generating a clean image, and are merely used to write a joint prior $p_X$ that factorizes across overlapping patches.


\subsection{Identification as Analysis Formulation}
The estimation criterion in \eqref{eq:analysis} clearly falls in the analysis-type category \cite{EladMilanfar}, since  it considers the image itself as the object to be estimated and as the argument of the prior. However, as a generative model for clean images, its meaning is not very clear, since  it is not a trivial task to obtain samples from this distribution.

To obtain a more compact notation, let $\bP: \mathbb{R}^{N} \rightarrow \mathbb{R}^{M n}$  be the operator (an $M n\times N$ matrix) that extracts the  set of $M$ patches, \textit{i.e.}, 
\begin{equation}
\bP \bx = \begin{bmatrix}
\bP_1 \bx \\
\vdots \\
\bP_M \bx
\end{bmatrix}\in \mathbb{R}^{Mn}.
\end{equation}
The prior $p_X$  may then be written as 
\begin{equation}
p_X(\bx) \propto p_V(\bP\bx), 
\end{equation}
where $p_V$ denotes a density defined in $\mathbb{R}^{Mn}$ according to 
\begin{equation}
p_V (\bv ) = \prod_{m=1}^{M} f(\bv_m).
\end{equation}
With this notation, the MAP denoising problem \eqref{eq:analysis} can be written as 
\begin{equation}
\hat{\bx}_{\mbox{\scriptsize MAP-A}} \in \arg\min_{\bx} \frac{1}{2\sigma^2}\|\bx - \by\|_2^2 -\log p_V (\bP \bx ) , \label{eq:patch-analysis2}
\end{equation}
which clearly reveals the analysis nature of this formulation (see \eqref{eq:analysis_general}).

\subsection{Patch~level Models}
We stress that the analysis/synthesis dichotomy  addressed in this paper concerns the way in which an image relates to its patches, not the way the patches themselves are modelled. In fact, the patch-analysis formulation just reviewed is compatible with a synthesis patch model, \textit{e.g.}, one that models each patch $\bP_m \bx$ as linear combination of elements of some dictionary $\bD$, with coefficients $\bgamma_m$ equipped with some prior $p_{\Gamma}$ (for example, a sparsity-inducing prior, as in \cite{SulamElad}). Using the half quadratic splitting approach, the formulation becomes
\begin{eqnarray}
\hat{\bx}_{\mbox{\scriptsize MAP-A}}  & \displaystyle \!\!\!\!\! \in  \arg \min_{\bx} \min_{\bgamma} & \!\! \frac{1}{\sigma^2} \|\by - \bx\|_2^2  + \beta \sum_{m=1}^M \|\bD\bgamma_m - \bP_m\bx\|_2^2 \nonumber \\ & & - 2 \sum _{m=1}^M \log p_{\Gamma}(\bgamma_m).\label{eq:EPLL2}
\end{eqnarray}
where $\bgamma = (\bgamma_m,\; m=1,...,M)$. Naturally, the patch-analysis formulation is also compatible with an analysis patch model \cite{Chen}, by using a patch prior of the form $f = \phi\circ \bA$, \textit{i.e.}, $f(\bv) = \phi(\bA\bv)$, for some function $\phi:\mathbb{R}^{s}\rightarrow \mathbb{R}_+$ and matrix $\bA\in\mathbb{R}^{s\times n}$.

To identify the analysis/synthesis dichotomy in terms of how the patches are related to the underlying image, not in terms of how the patches are modelled, we refer to the formulation reviewed in this section as \textit{patch-analysis} and to its synthesis counterpart (to be introduced in the next section) as \textit{patch-synthesis}.

\section{Patch-Synthesis Formulation}\label{sec:patch_s}
We now present the patch-synthesis formulation, which can be summarized as follows: the clean image is generated by additively combining a collection of patches; the patches themselves follow some probabilistic model, but are a priori mutually independent. 

Consider a collection of patches $\{\bz_m \in \mathbb{R}^n,\; m=1,...,M\}$ and let image $\bx$ be synthesized  from these patches by combining them additively according to 
\begin{equation}
\bx = \sum_{m=1}^M \bQ_m \, \bz_m, \label{eq:synth1}
\end{equation}
where matrices $\bQ_m \in \mathbb{R}^{N\times n}$ are such that they average the  values in the several patches that contribute to a given pixel of $\bx$. 

A simple 1D example will help clarify the structure of the $\bQ_m$ matrices. Consider that $\bx = [x_1,x_2,x_3,x_4]^T$  is produced by combining all the consecutive 2-element patches (with periodic boundary conditions), which correspond to the subsets of components $\{1,2\}$, $\{2,3\}$, $\{3,4\},$ and $\{4,1\}$. That is, $M=4$ and
\begin{eqnarray}
&\bQ_1  = \begin{bmatrix}
1/2 & 0 \\ 0 & 1/2 \\ 0 & 0 \\ 0 & 0
\end{bmatrix}, 
& \bQ_2  = \begin{bmatrix}
0 & 0 \\ 1/2 & 0 \\ 0 & 1/2 \\ 0 & 0
\end{bmatrix}
, \label{eq_example1} \\
& \bQ_3  = \begin{bmatrix}
0 & 0 \\ 0 & 0 \\ 1/2 & 0 \\ 0 & 1/2
\end{bmatrix}, 
& \bQ_4  = \begin{bmatrix}
0 & 1/2 \\ 0 & 0 \\ 0 & 0 \\ 1/2 & 0
\end{bmatrix}. \label{eq_example2}
\end{eqnarray}

Stacking all the patches in vector $\bz \in \mathbb{R}^{Mn}$ and considering a matrix $\bQ = [\bQ_1 \cdots \bQ_M] \in \mathbb{R}^{N\times(Mn)}$, the synthesis expression in \eqref{eq:synth1} can be written compactly as 
\begin{equation}
\bx = \bQ \bz.\label{eq:master_synth}
\end{equation}

With the patches modelled as independent and identically distributed samples of some patch-wise pdf $g$, their joint log-prior is
\begin{equation}
\log p_Z(\bz) = \sum_{m=1}^M \log g(\bz_m). \label{eq:synth-patches}
\end{equation}

Notice that obtaining samples from \eqref{eq:synth-patches} is as simple as obtaining samples from $g$ itself. Consequently, generating image samples under this synthesis model simply corresponds to generating samples from \eqref{eq:synth-patches} and then multiplying them by $\bQ$. This is in contrast with the patch-analysis prior \eqref{eq:PoE}, where, even if $f$ is a valid pdf, it is not trivial to obtain samples from $p_X$.

The resulting MAP denoising criterion can now be written as 
\begin{equation}
\hat{\bx}_{\mbox{\scriptsize MAP-S}} = \bQ \hat{\bz},\;\mbox{where}\; \hat{\bz} \in \arg\min_{\bz}  \frac{1}{2\sigma^2}\|\bQ\bz - \by\|_2^2 - \log p_Z( \bz ).\label{eq:patch-synthesis2}
\end{equation}

As the patch-analysis model, the patch-synthesis formulation that we have just presented is compatible with both analysis and synthesis priors for the patches, and of course with any valid pdf for vectors in $\mathbb{R}^n$. An analysis formulation simply amounts to  choosing a patch prior of the form $g = \psi\circ \bB$, \textit{i.e.}, $g(\bz_m) = \psi(\bB\bz_m)$, for some function $\psi:\mathbb{R}^{s}\rightarrow \mathbb{R}_+$ and matrix $\bB\in\mathbb{R}^{s\times n}$.

In a synthesis formulation, each patch $\bz_m$ is synthesized using some dictionary $\bD$ as $\bz_m = \bD\bgamma_m$, and the $\bgamma_m$ follow some prior $p_{\Gamma}$; stacking all the patch coefficients in vector $\bgamma$, we can write $\bz = \bDelta\bgamma$ (where $\bDelta$ is a block-diagonal matrix with $M$ replicas of $\bD$), thus the MAP denoising criterion  becomes
\begin{equation}
\hat{\bx}_{\mbox{\scriptsize MAP-S}} = \bQ  \bDelta \hat{\bgamma},
\end{equation}
where
\begin{equation}
 \hat{\bgamma} \in \arg\min_{\bgamma}  \frac{1}{2\sigma^2}\|\bQ \bDelta \bgamma - \by\|_2^2 - \log q_{\Gamma}( \bgamma ).
\end{equation}

\section{Relationship Between the Analysis and Synthesis Formulations}\label{sec:rel}
Leaving aside for now the choice of the patch priors, let us focus on the relationship between formulations \eqref{eq:patch-analysis2} and \eqref{eq:patch-synthesis2}. The key observation underlying the relationship between these two formulations is that (assuming the patch structure in both formulations is the same) 
\begin{equation}
\bQ \, \bP = \bI,\label{eq:pseudo}
\end{equation}
where $\bI$ denotes the identity matrix, but in general
\begin{equation}
\bP \, \bQ \neq \bI.\label{eq:pseudo2}
\end{equation} 
In other words, $\bQ$ is a left pseudo-inverse of $\bP$. To prove \eqref{eq:pseudo}, simply notice that if a collection of patches is extracted from some image and then these patches are used to synthesize an image by averaging the overlapping pixels, an identical image is obtained. Of course, the converse is not true, in general: if an image is synthesized by averaging the overlapping pixels of a collection of patches, and then patches are extracted from the synthesized image, there is no guarantee that these patches are equal to the original ones, thus proving \eqref{eq:pseudo2}. In the trivial and uninteresting cases where the patches are singletons, or non-overlapping, we would have $\bP \, \bQ = \bI$.

As shown in \cite{EladMilanfar}, given the analysis formulation \eqref{eq:patch-analysis2}, an equivalent synthesis formulation is
\begin{equation}
\begin{split}
\hat{\bx}_{\mbox{\scriptsize MAP-A}} = \bQ \hat{\bz},\; \mbox{where}\;  \hat{\bz} \in & \arg\min_{\bz}   \frac{1}{2\sigma^2}\|\bQ\bz - \by\|_2^2 - \log p_V( \bz ),  \\ 
& \st \bz\in\range(\bP),\label{eq:patch-synthesis3}
\end{split}
\end{equation}
where the constraint $\bz\in\range(\bP)$ enforces $\bz$ to be in the subspace spanned by the columns of $\bP$. Notice that this constraint corresponds to having a collection of patches extracted from some image, \textit{i.e.}, it forces the patches to agree on the value of each shared pixel. 
Since \eqref{eq:patch-synthesis2} does not enforce this constraint, it is not, in general, equivalent to the patch-analysis formulation \eqref{eq:patch-analysis2}.

\section{ADMM for Patch-Synthesis Denoising}\label{sec:admm}
In this section, we derive an instance of ADMM to deal with \eqref{eq:patch-synthesis2}. The first step is to rewrite it as a constrained problem,
\begin{equation}
\begin{split}
& \min_{\bz,\bu}  \;  \frac{1}{2\sigma^2}\|\bQ\bz - \by\|_2^2 + \xi( \bu ),  \\ 
& \st \; \bz =  \bu, \label{eq:split}
\end{split}
\end{equation}
where 
\begin{equation}
\xi(\bu) =  - \log p_Z(\bu) = \sum_{m=1}^M \underbrace{(-\log g(\bu_m) )}_{\xi_m (\bu_m)}\label{eq:xi}
\end{equation}
 is the negative log-prior, or regularizer.  ADMM for this problem takes the form
\begin{eqnarray*}
\bz^{(t+1)} & = & \argmin_{\bz}  \; \frac{1}{2\sigma^2}\|\bQ\bz - \by\|_2^2  + \frac{\rho}{2}\| \bz - \bu^{(t)} - \bd^{(t)} \|_2^2 \\
\bu^{(t+1)} & = & \argmin_{\bu} \; \xi(\bu) + \frac{\rho}{2}\| \bz^{(t+1)} - \bu - \bd^{(t)} \|_2^2 \\
\bd^{(t+1)} & = & \bd^{(t)} + \bz^{(t+1)} - \bu^{(t+1)}
\end{eqnarray*}
The update equation for $\bu^{(t+1)}$ is a denoising step. Due to the separability of the squared $\ell_2$ norm and of $\xi$ (see \eqref{eq:xi}), this can be separately solved with respect to each patch: for $m=1,...,M$,
\begin{equation}
\bu_m^{(t+1)}  =  \argmin_{\bu_m } \; \xi_m(\bu_m) + \frac{\rho}{2}\| \bz_m^{(t+1)} - \bu_m - \bd_m^{(t)} \|_2^2,\label{eq:u_update}
\end{equation}
which is simply the proximity operator of $(1/\rho)\xi_m $ \cite{BauschkeCombettes}, computed at $\bd_m^{(t)} - \bz_m^{(t+1)}$. In a Bayesian viewpoint, \eqref{eq:u_update} corresponds to obtaining the MAP estimate of $\bu_m$ from observations $\bd_m^{(t)} - \bz_m^{(t+1)}$, assuming additive white Gaussian noise of variance $1/\rho$ and a negative log-prior $\xi_m$. If the $\xi_m$ are convex, this MAP estimate is unique, due to the strict convexity of the quadratic term in \eqref{eq:u_update}.

Computing $\bz^{(t+1)}$ corresponds to solving an unconstrained quadratic problem, the solution being
\begin{equation}
\bz^{(t+1)} = \bigl(\bQ^T\bQ + \sigma^2\rho\bI\bigr)^{-1} \bigl( \bQ^T\by + \sigma^2 \rho\,  (\bu^{(t)} + \bd^{(t)})\bigr).\label{eq:updatez}
\end{equation}
The bottleneck in this update equation seems to be the matrix inversion, since $\bQ^T\bQ$ is a huge $(Mn)\times(Mn)$ matrix. However, this inversion can be solved very efficiently by resorting to the Sherman-Morrison-Woodbury matrix inversion formula. In fact,
\begin{equation}
\bigl(\bQ^T\bQ + \sigma^2\rho\, \bI\bigr)^{-1} = \frac{1}{\sigma^2\rho} \Bigl( \bI - \bQ^T (\sigma^2 \rho\, \bI + \bQ\, \bQ^T)^{-1} \bQ \Bigr),
\end{equation}
where matrix $(\sigma^2 \rho\, \bI + \bQ\, \bQ^T)$ is diagonal, thus its inversion is trivial. 
To prove that $\bQ\bQ^T$ is diagonal, recall that $\bQ = \begin{bmatrix}
\bQ_1 \; \bQ_2 \; \cdots \bQ_M
\end{bmatrix}$, thus
\begin{equation}
\bQ\bQ^T =  \sum_{m=1}^M \bQ_m\bQ_m^T.
\end{equation}
The element $(i,j)$ of matrix $\bQ_m\bQ_m^T \in \mathbb{R}^{N\times N}$ is the inner product between the $i$-th and the $j$-th rows of $\bQ_m$. Since each pixel in the $m$-th patch contributes to one and only one pixel in the synthesized image, the rows of $\bQ_m$ have disjoint support, thus $(i\neq j) \Rightarrow (\bQ_m\bQ_m^T)_{i,j}=0$, that is, $\bQ_m\bQ_m^T$ has no non-zero elements outside of its main diagonal, thus is a diagonal matrix. Finally, since $\bQ\bQ^T$ is a sum of diagonal matrices, it is a diagonal matrix.

It is also easy to obtain explicitly the elements of the diagonal of $\bQ\bQ^T$. The diagonal elements $(\bQ_m\bQ_m^T)_{i,i}$ are given by 
\begin{equation}
(\bQ_m\bQ_m^T)_{i,i} = \sum_{j=1}^n ((\bQ_m)_{i,j})^2. \label{eq:diagsQmQ}
\end{equation}
If the $m$-th patch contributes to pixel $i$, the sum in \eqref{eq:diagsQmQ} contains exactly one non-zero term, thus it is equal to the square of the weight with which the $m$-th patch contributes to the synthesis of pixel $i$. If the $m$-th patch does not contributes to pixel $i$, then $(\bQ_m\bQ_m^T)_{i,i} = 0$. Finally, since the weight with which each patch element contributes to each pixel equals the inverse of the number of patches that contributes to that pixel, $(\bQ\bQ^T)_{i,i}$ is equal to the inverse of the number of patches that contribute to pixel $i$.
Referring to the example in \eqref{eq_example1}--\eqref{eq_example2}, we have simply $\bQ\bQ^T = (1/2)\bI$, because each pixel is synthesized from two patches.

Finally, letting $\bq = \mbox{diag}\bigl( (\sigma^2\rho \bI  + \bQ\bQ^T)^{-1}\bigr)$, and denoting $\bs^{(t)} =  \bQ^T\by + \sigma^2 \rho (\bu^{(t)} + \bd^{(t)})$, we can write the update equation \eqref{eq:updatez} as
\begin{equation}
\bz^{(t+1)} = \frac{1}{\sigma^2\rho} \Bigl(\bs^{(t)}  - \bQ^T \bigl( \bq \odot (\bQ \bs^{(t)})\bigr)\Bigr),\label{eq:updatez2}
\end{equation}
where $\odot$ denotes element-wise product between two vectors. The leading cost of this update is that of the matrix-vector products involving $\bQ$ and $\bQ^T$, which is $O(MNn)$; all the other operations in \eqref{eq:updatez} have lower computational cost.

\section{Conclusions and Future Work}\label{sec:conclusion}
In this paper, we have revisited patch-based image priors under the light of the synthesis vs analysis dichotomy. After showing that the classical patch-based image models (namely the EPLL) corresponds to an analysis formulation, we have proposed a patch-synthesis formulation, and analyzed its relationship with the analysis formulation, showing that they are, in general, not equivalent. Finally, we have shown how to address image denoising under the proposed formulation, via an ADMM algorithm.

We stress again that the purpose of this paper is not to introduce a new particular image prior, but a general patch-based synthesis formulation/framework, which (to the best of our knowledge) was missing from the literature on patch-based image processing, and which can be instantiated with many different patch models/priors. For this reason, we have abstained from presenting experimental results; these would critically depend on the choice and estimation of a particular patch model, which is not the focus of this paper.

Ongoing work includes the development of efficient algorithms for learning patch models under the assumption that they will be used in a synthesis formulation.

\end{document}